\documentclass[a4paper, 10pt, conference]{IEEEtran}
\usepackage{cite}
\usepackage{amsmath,amssymb,amsfonts}
\usepackage{algorithmic}
\usepackage{graphicx}
\usepackage{textcomp}
\usepackage{xcolor}
\def\BibTeX{{\rm B\kern-.05em{\sc i\kern-.025em b}\kern-.08em
T\kern-.1667em\lower.7ex\hbox{E}\kern-.125emX}}
\begin{document}

\title{Advances in Preference-based Reinforcement Learning: A Review\
}

\author{\IEEEauthorblockN{1\textsuperscript{st} Youssef Abdelkareem}
\IEEEauthorblockA{\textit{Electrical and Computer Engineering} \\
\textit{University of Waterloo}\\
Waterloo, Canada \\
yafathi@uwaterloo.ca}
\and
\IEEEauthorblockN{2\textsuperscript{nd} Shady Shehata}
\IEEEauthorblockA{\textit{Electrical and Computer Engineering} \\
\textit{University of Waterloo}\\
Waterloo, Canada \\
sshehata@uwaterloo.ca}
\and
\IEEEauthorblockN{3\textsuperscript{rd} Fakhri Karray}
\IEEEauthorblockA{\textit{Electrical and Computer Engineering} \\
\textit{University of Waterloo}\\
Waterloo, Canada \\
karray@uwaterloo.ca}
}

\maketitle

\begin{abstract}
Reinforcement Learning (RL) algorithms suffer from the dependency on accurately engineered reward functions to properly guide the learning agents to do the required tasks. Preference-based reinforcement learning (PbRL) addresses that by utilizing human preferences as feedback from the experts instead of numeric rewards. Due to its promising advantage over traditional RL, PbRL has gained more focus in recent years with many significant advances. In this survey, we present a unified PbRL framework to include the newly emerging approaches that improve the scalability and efficiency of PbRL. In addition, we give a detailed overview of the theoretical guarantees and benchmarking work done in the field, while presenting its recent applications in complex real-world tasks. Lastly, we go over the limitations of the current approaches and the proposed future research directions.
\end{abstract}

\section{Introduction}
Reinforcement learning (RL) is a sub-field of machine learning that has been widely implemented in many applications such as robot control \cite{robotics_survey}, games \cite{survey_games}, and medical domains \cite{healthcare}. A learning agent continuously interacts with an environment by doing actions and receives feedback signals (rewards) with the target of maximizing the cumulative reward by the end of the interaction phase. The reward received is assumed to be numerically generated from a specific reward function. The main issue that arises is the high sensitivity of the performance to the design of the reward function. Specifically, one of the challenges in reward engineering is called Reward Hacking which occurs when the agent comes up with ways to maximize the cumulative rewards without executing the intended task \cite{reward_hacking}. 
Reward Shaping is also another problem that involves finding the optimal balance between providing rewards that direct the agent to the final goal (extrinsic motivation) while guiding them to also do the intended task at the same time (intrinsic motivation) \cite{reward_shaping}.

\begin{figure*}[htbp]
\centerline{\includegraphics[width=1.0\textwidth]{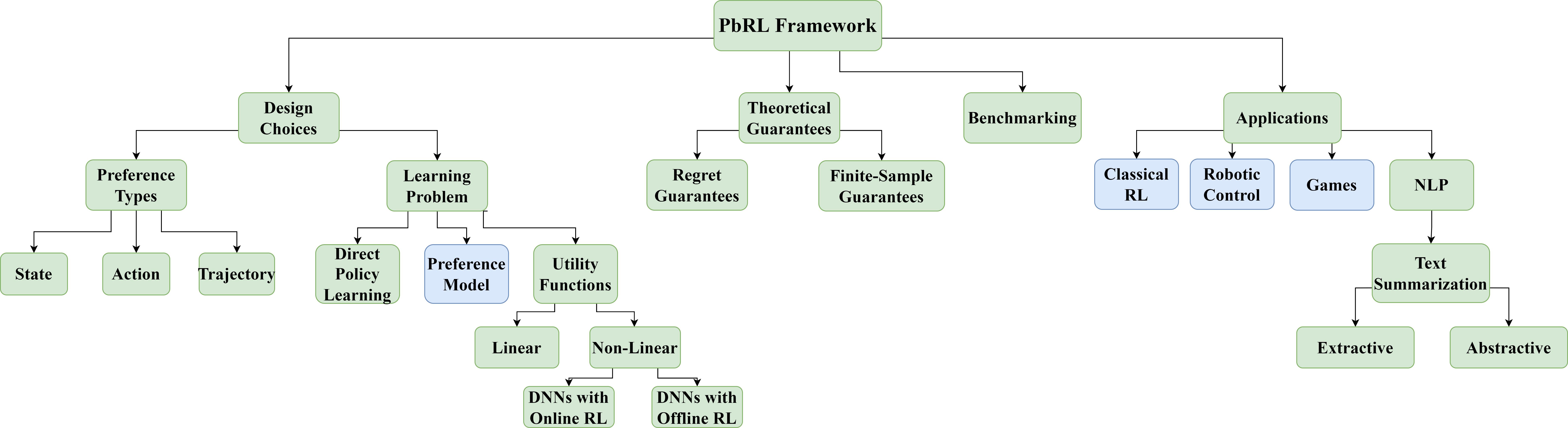}}
\caption{Hierarchical structure of our PbRL framework. Topics in green are surveyed in this work, while the ones in blue are not within our scope. "DNNs" denotes deep neural networks.}
\label{fig:my_label}
\end{figure*}

The field of preference-based reinforcement learning (PbRL) promises a solution to the aforementioned problems. It revolves around providing the agent with non-numeric reward signals in the form of pairwise preferences rather than absolute rewards. This shift in approach broadens the scope of RL algorithms to non-expert users where accurate reward engineering is no longer required. In this work, we present a coherent framework for PbRL, summarized in Figure \ref{fig:my_label}. Our framework is an extension of the one proposed by \cite{pbrl_survey} where we include the most recent advances in PbRL. 

\break
We start with a formal definition of the problem in Section \ref{pre_knowledge}, followed by the different design choices in Section \ref{design_choice}.
We then go over the PbRL algorithms with theoretical guarantees in Section \ref{theoretical} and the available benchmarking frameworks in Section \ref{benchmarking}. Applications of PbRL in the Natural Language Processing (NLP) domain are presented in Section \ref{applications}. Lastly, we conclude by analyzing the shortcomings of the surveyed methods and propose future research directions in Section \ref{analysis}.


\section{Problem Formulation}
\label{pre_knowledge}
In PbRL algorithms, we are trying to solve the traditional RL problems using preferences between pairs of states, actions, or trajectories rather than absolute numerical rewards. 
The MDP for preferences (MDPP) \cite{pbrl_survey} is represented as a sextuple $(S,A,\mu,\delta,\gamma,\rho)$. Similar to the original MDP, $S$ and $A$ denote the state and action spaces that could be either discrete, with sizes $|S|$ and $|A|$, or continuous. $\delta(s'|s,a)$ represents a stochastic state transition model, while $\gamma$ is the discount factor $\in [0,1)$, $\mu(s)$ is the initial state distribution and $h$ is the horizon length in finite-horizon settings. Trajectories ($\tau$) define a sequence of state-action pairs and $\pi(a|s)$ represents the policy. The difference between MDP and MDPP is that a preference relation over trajectories $\tau_1 \succ \tau_2$, where $\tau_1$ is more preferred than $\tau_2$, is received by the agent instead of the numeric reward signal $r(s,a)$. $\rho(\tau_1 \succ \tau_2)$ denotes the probability that a preference relation holds. 
Regarding the objective, we would like the agent to learn an optimal policy $\pi^*$ that generates trajectories that satisfy the set of all preference relations $\zeta$ received from the expert during training. 
The objective of PbRL for a single preference relation assumes that the optimal policy is the one that maximizes the difference between the probabilities of obtaining the more preferred trajectory (dominating) and the less preferred one (dominated).

\section{Design Choices}
\label{design_choice}
\subsection{Preference Type Design Choices}
\label{action-pref}

In the literature, the types of preferences that are received from experts could be divided into action, state, and trajectory preferences. 

\subsubsection{Action Preferences}
Action preferences reduce the preference relation to the comparison of a pair of actions for the same state, where $a_1 \succ_s a_2$ denotes that action $a_1$ is more preferred than action $a_2$ for state $s$. 

\noindent Utilizing action preferences that optimize short-term rewards would be hard as they're only valid for a given state. \cite{Frnkranz2012PreferencebasedRL} uses action-based preferences that deal with long-term optimality and evaluate the relation $a_1 \succ_s a_2$ by doing a roll-out for trajectories starting by state $s$, doing action $a$, and following the estimate of the policy to get the expected returns of each action. 
Action preferences with long-term optimality are considered demanding for experts since the long-term outcome should be known.
\subsubsection{State Preferences}
Regarding state-preferences, a relation $s_1 \succ s_2$ indicates that $s_1$ is preferred over $s_2$. Since those relations correspond to segments of the state space, they give more information compared to action preferences and are less demanding to the expert since no comparisons between actions are needed. \cite{Wirth2012FirstST, Wirth2015OnLF} follow long-term optimality by proposing that selecting the most preferred successor state for every state could be a viable solution in their long-term state-based setting. \cite{Zucker2010AnOA} uses short-term state preferences and mitigates their issues by trying to estimate a cost function for every state using the support vector ranking approach \cite{Herbrich1999SupportVL}. 
\subsubsection{Trajectory Preferences}
The most common type of preference relations are trajectory preferences where $\tau_1 \succ \tau_2$ indicates that trajectory $\tau_1$ dominates over $\tau_2$. Such preferences are desirable since they can be easily evaluated by experts by assessing the full trajectories and their results. Most of the methods presented in the rest of this paper use trajectory preferences. A general challenge is relating the final preference over trajectories to their most relevant states and actions.

\subsection{Learning Problem Design Choices}
\label{lp_design_choice}
There have been several proposed approaches to use the preference feedback received from the expert to optimize the policy. Our main focus will be on approaches that directly learn a policy distribution, or estimate a utility function. Other approaches like \cite{Frnkranz2012PreferencebasedRL} learn a preference model, usually modeled as a classifier, to predict whether a preference relation holds between two actions for a given state.
\subsubsection{Learning a Policy}

Some methods directly derive an estimation of the policy. \cite{Wilson2012ABA} learns a policy distribution using a Bayesian likelihood function. Specifically, they utilize the posterior distribution of policies $P(\pi|\zeta)$ given the preference relations to sample two parameterized policies which are used to generate two full trajectories $\tau_1, \tau_2$ starting from the same state. The sampled trajectories are transformed into a trajectory preference relation $(\tau_1 \succ \tau_2)$ and added to the existing buffer storing all the preference relations. The posterior policy distribution $P(\pi|\zeta)$ is represented with Bayes theorem through the multiplication of the prior distribution of the policy $P(\pi)$ with the likelihood of all the trajectory preferences $P(\tau_1 \succ \tau_2|\pi) \in \zeta$ and approximated using Markov Chain Monte Carlo Simulation \cite{Andrieu2004AnIT}. The downside is that the likelihood is modeled in terms of the euclidean distance between the policy-realized trajectories which constrains the algorithm to perform properly on low-dimensional continuous state spaces only. \cite{BusaFekete2013PreferencebasedED,BusaFekete2014PreferencebasedRL} mitigate this issue by not requiring a distance function in their objective. They make a direct policy comparison by presenting trajectory preference queries comparing trajectories $\tau^{\pi_1}, \tau^{\pi_2}$ generated by the two policies $\pi_1, \pi_2$. 

\noindent The sets of all preference relations and policies are used to generate a ranking between policies returning the highest-ranked policies as the optimal ones. However, their method requires a higher number of preference queries than \cite{Wilson2012ABA}, as their preferences are non-reusable, making it not feedback-efficient.

\subsubsection{Learning a utility function}
\label{utility_func}
Learning a policy directly can be highly sample-inefficient, therefore, some methods try to estimate a surrogate utility function $U(x)$, where $x$ denotes trajectories or state-action pairs, to extract more information from the preferences. This utility function is analogous to the reward function seen in RL, however, they are not directly related since the definition of what is optimal is dependent on the views of the expert giving the preference feedback. There are two types of utility functions used in the literature which are \textit{linear} and \textit{non-linear}. For both types, trajectories and state-action pairs are assumed to have a feature vector representation denoted by $\psi(\tau)$ and $\varphi(s, a)$ respectively. The general objective is to obtain the optimal policy by maximizing a \textit{link function} $d$ that represents the difference between the utilities of the dominating and dominated preference relation terms.

\hspace{1em } \textit{a) Linear utility functions:}
\label{linear_utility_func}
This type of utility is formulated for trajectories in terms of an unknown weight vector $\theta$ yielding $U(\tau) = \theta^T \psi(\tau)$. Methods that use such type of linear utility formulate the link function for a trajectory preference relation as follows $d(\theta,\tau_1 \succ \tau_2)=\theta^T(\psi(\tau_1)-\psi(\tau_2))$ and find the optimal $\theta$ that maximizes the link function for all preference relations. The optimization depends on choosing a proper loss function which differs based on the proposed methods. \cite{Akrour2012APRILAP} incorporated the hinge loss that is approximated using the ranking SVM method following \cite{Herbrich2001BayesPM} and estimated the hand-coded feature representations $\psi(\tau)$ for the trajectories using $\epsilon$-means clustering \cite{Duda1973PatternCA}.
\cite{Schoenauer2014ProgrammingBF} proposes an enhancement by accounting for the inaccuracies of the expert through the introduction of a piece-wise loss. The loss represents the inaccuracies using a ridge noise model controlled by a dynamically changing hyper-parameter to consider preferences that change over time. 

\hspace{1 em } \textit{b) Non-linear utility functions:} 
\label{non_linear_utility_func}
There are different ways to introduce non-linearity in the utility functions. 
Our focus is on recent methods that utilize deep neural networks (DNNs) as a non-linear representation for the utility. Those methods can be categorized based on using online or offline RL algorithms in their formulation.

\hspace{2 em } \textit{b.1) DNNs with Online RL:} \cite{Christiano2017DeepRL} is the first method to represent the parameters $\theta$ of the utility function $U_\theta(s,a)$ with a DNN. Due to their high representational capacity, DNNs opened the door to experiment on more challenging robotic \cite{Todorov2012MuJoCoAP} and Atari \cite{Bellemare2015TheAL} tasks.
In addition, the experts are queried with trajectory segments ($\sigma$) instead of full trajectories ($\tau$) which reduces the effort done by them compared to \cite{Akrour2012APRILAP}.
The utility of the segments are sum-decomposable in terms of the non-linear state-action utilities yielding $U_\theta(\sigma)=\sum_i U_\theta(s_i,a_i)$. 

\break
\noindent The link function for a trajectory segment preference relation $d(\theta,\sigma_1 \succ \sigma_2$) is modeled as a probability function following the Bradley-Terry model \cite{Bradley1952RankAO} and optimized using the cross-entropy loss with the labels being the preference feedback received from the experts.
\begin{equation}
\label{eq:christiano}
d(\theta,\sigma_1 \succ \sigma_2) = \frac{e^{U_\theta(\sigma_1)}}{e^{U_\theta(\sigma_1)}+e^{U_\theta(\sigma_2)}}
\end{equation}

One could look at the approach of \cite{Christiano2017DeepRL} as a model-free RL algorithm, where the state-transition dynamics are unknown and the reward model corresponds to the utility function $U_\theta(s, a)$ which mimics the reward received from the environment. Based on that, they utilize online policy gradient approaches like A2C \cite{Mnih2016AsynchronousMF} and TRPO \cite{Schulman2015TrustRP} to optimize the policies while using the learned utility as an estimate of the reward received from the environment with each interaction. This online learning setting needs many interactions with the environment and numerous preference queries to reach the optimal policy. \cite{Cao2021WeakHP} focus on adding a supervised human-preference estimator that learns to mimic the human preferences to decrease the number of preference queries presented to the expert by depending on the estimated preferences for some samples. Furthermore, they allow the expert feedback to be any continuous value in the range [0,1] to deal with weak preference relations between trajectory segments. Both contributions enhanced the performance and efficiency compared to \cite{Christiano2017DeepRL}.

\hspace{2em } \textit{b.2) DNNs with Offline RL}. Offline methods train on examples saved in a buffer to reduce the interactions with the environment. \cite{Lee2021PEBBLEFI} proposed utilizing the famous Soft Actor-Critic (SAC) offline RL algorithm \cite{Haarnoja2018SoftAO} with the same utility function settings used by \cite{Christiano2017DeepRL} to handle the sample and feedback inefficiency problems. To do that, they had to account for the dynamically changing estimate of the utility function by relabelling all the agent's past sequences with the corresponding new utility values whenever the utility function is updated with new preferences. Additionally, they tackled the limited diversity in the initial preference queries given to the expert, which happens due to the random initialization of the policy, by including an unsupervised pre-training step. The main limitation of this approach is the high cost of relabelling all the preferences stored in the buffer. \cite{park2022surf} build on the offline learning approach of \cite{Lee2021PEBBLEFI} by introducing a pseudo-labeling procedure on the existing unlabeled preference relations stored in the offline dataset. 

\noindent Specifically, an unlabeled preference relation $\sigma_1 \succ \sigma_2$ is predicted to hold if the probability estimated with the link function from \eqref{eq:christiano} exceeds 50\% and doesn't hold otherwise. Only confident labels with high probability are used. Moreover, they introduce the first usage of data augmentation approaches in PbRL to enhance the regularization and feedback efficiency even more. Such augmentation is done by randomly cropping labeled segments under the assumption that the preference label should be consistent with the re-scale and shift of the individual segments.

\section{Theoretical Guarantees}
\label{theoretical}
There is a current research target aiming to develop novel PbRL algorithms that are tractable for theoretical analysis. Those algorithms focus on reaching either regret or finite-sample guarantees which will be discussed in detail in this section.
\subsection{Regret Guarantees}
Regret denotes the difference between the current expectation of total rewards and the maximum rewards generated by the optimal policy. 
Providing theoretical guarantees on the bounds of the regret has been an important aim for various RL and bandit approaches \cite{Russo2014LearningTO, Domingues2020RegretBF}. DPS \cite{Novoseller2020DuelingPS} is the first paper to propose a PbRL algorithm with solid regret guarantees. They base their algorithm on the Thompson Sampling algorithm \cite{Thompson1933ONTL} while formulating Bayesian regret bounds by borrowing the information-theoretic concepts from \cite{Russo2016AnIA}. DPS uses trajectory preferences within a model-based approach. They assume having a non-linear utility function $U(s,a)$ for state-action pairs to estimate the reward model. Both the reward and transition dynamics models are represented as Bayesian posterior distributions. Similar to \cite{Christiano2017DeepRL}, the trajectory utilities are a summation of state-action utilities yielding $U(\tau)=\sum_i U(s_i,a_i)$. The link function is linear with $d(\theta,\tau_1 \succ \tau_2)=U(\tau_1)-U(\tau_2)$ where $\theta$ represents the parameters of the distribution of the utility function. Their algorithm samples two distinct pairs of transition and utility models from their distributions and applies value iteration to yield the two corresponding policies. 
The transition and utility distributions are then updated using the queried preferences history present in a buffer. 

\break
\noindent By proving the asymptotic convergence of the transition and reward models, they were able to reach an asymptotic sublinear regret rate of $|S|\sqrt{2|A|Nh\log|A|}$, where $N$ is the number of iterations of the algorithm.The main limitation of \cite{Novoseller2020DuelingPS} is the exponential complexity in the time horizon $h$ for the asymptotic convergence of the reward and transition models. \cite{Pacchiano2021DuelingRR} proposed a more general PbRL algorithm with regret guarantees by assuming that the underlying utility function represents non-Markovian rewards. The utility function is linear in terms of the trajectory features with dimension $d$, such that $U_\theta(\tau)=\theta^T\psi(\tau)$. They utilize the same link function in \eqref{eq:christiano} in terms of full trajectories and add an extra L2 regularization term on $\theta$. The foundation of their approach is based on bounding $\theta$ under a parameter $Q$ with the rescaling concepts used in \cite{Faury2020ImprovedOA}. Consequently, they propose two algorithms that assume having known and unknown transition models achieving near-optimal regret bounds of $O'(Q d \log (N/\delta)\sqrt{N})$, with a probability of at least $1-\delta$, and $O'((\sqrt{d}+h^2+|S|)\sqrt{dN}+\sqrt{|S||A|Nh})$, respectively. The notation $O'$ hides any logarithmic factors in the variables.
\subsection{Finite-Sample Guarantees}

\cite{Xu2020PreferencebasedRL} is the only PbRL algorithm in the literature to derive finite-sample guarantees in terms of the number of preferences queries and the number of interactions with the environment (number of steps). Trajectory preferences are used under the observation that those preferences need to be noisy to derive a unique optimal policy. The target of the algorithm is to efficiently obtain an $\epsilon$-optimal policy. They utilize black-box PAC-Dueling Bandits (P-DB) algorithms like \cite{Yue2011BeatTM} and \cite{Falahatgar2017MaximumSA} to make policy comparisons based on the collected trajectory preferences. Specifically, one of their proposed algorithms (PEPS) explores the state space by synthesizing a reward function, similar to reward-free RL \cite{Jin2020RewardFreeEF}, and optimizes it using a tabular RL algorithm (EULER; \cite{Zanette2019TighterPR}). The P-DB algorithm then generates action queries during learning which are transformed into trajectory preference queries by rolling out the trajectories with the current policy estimated with EULER. If the P-DB algorithm requires no target accuracy in advance, the PEPS method reaches an $\epsilon$-optimal policy with a step complexity of $O(\frac{h^2|S|^2|A|\iota}{\epsilon^2}+\frac{|S|^4|A|h^3\iota^3}{\epsilon})$ and a preference query complexity of $O(\frac{h|S|^2|A|\iota}{\epsilon^2})$, where $\iota$ denotes the log factors. The main limitations of this approach are the sub-optimal sample complexity and the dependency on the guarantees of the underlying P-DB algorithm which only hold under some restrictions on the structure of the preferences.
\section{Benchmarking}
\label{benchmarking}
There has been a large focus in the literature to create benchmarks for various RL domains such as Offline RL \cite{Gulcehre2020RLUA}, and Safe RL \cite{Achiam2019BenchmarkingSE}. \cite{Lee2021BPrefBP} is the first paper to propose a benchmark for the consistent evaluation of PbRL algorithms without relying on expensive human feedback. To achieve that, they simulate an expert providing preference based on the total sum of the ground-truth rewards while explicitly accounting for the human errors. 

\break
\noindent Modeling the stochasticity of the simulated expert preferences is achieved by following the link function formulation in \eqref{eq:christiano} to model the preference probability of segments in terms of ground-truth rewards yielding
\begin{equation}
\label{eq:benchmark}
P(\sigma_1 \succ \sigma_2) = \frac{e^{\beta{\sum_{i=0}^{T} \gamma^{T-i}r(s_i^1,a_i^1)}}}{e^{\beta{\sum_{i=0}^{T} \gamma^{T-i}r(s_i^1,a_i^1)}}+e^{\beta{\sum_{i=0}^{T} \gamma^{T-i}r(s_i^2,a_i^2)}}},
\end{equation}
where $\beta$ controls the degree of expert determinism and $\gamma$ is a discount factor used to model the short-sightedness by giving higher weights to recent rewards. In addition, incomparable queries are skipped if the total segment reward is less than a threshold. The simulated expert is allowed to make mistakes by flipping \eqref{eq:benchmark} randomly with probability $\epsilon$. 

\cite{Lee2021BPrefBP} quantitatively measures the performance of PbRL algorithms by normalizing the average predicted returns for the true reward. 
In their experiments, they focus on comparing the performance of the state-of-the-art deep PbRL algorithms with non-linear utility functions \cite{Christiano2017DeepRL, Lee2021PEBBLEFI} explained in Section \ref{non_linear_utility_func}. The tasks used have absolute-valued states and dense rewards and are taken from the DeepMind Control Suite \cite{Tassa2018DeepMindCS} and the Meta-world benchmark \cite{Yu2019MetaWorldAB}. The results of the experiments indicate that both \cite{Christiano2017DeepRL} and \cite{Lee2021PEBBLEFI} only perform well in cases where the expert doesn't make errors while the performance significantly degrades once the expert preferences become more stochastic. 

\section{Application Areas in NLP}
\label{applications}
PbRL algorithms have been used in practical applications that involve robot teaching tasks, board games, and others.
We refer the reader to a detailed overview of those application domains discussed in \cite{pbrl_survey}.
In this survey, we focus on the application of PbRL algorithms to the text summarization task in Natural Language Processing (NLP) which predicts a qualitative summary by extracting the important information in an input piece of text. Solving the task using preferences instead of rewards, like ROUGE scores, led to a higher correlation with the quality of the summaries as judged by humans \cite{Schluter2017TheLO}.
The two main categories of the task are \textit{extractive} and \textit{abstractive} summarization.

\subsection{Extractive summarization}Summaries of this type are built by extracting specific sentences from the original text. 
\cite{Gao2018APRILIL} is one of the initial approaches utilizing PbRL algorithms for the task. In their MDP formulation, the state is the summary made so far and actions correspond to possible sentences to be extracted. They follow a linear utility-based PbRL approach discussed in Section \ref{linear_utility_func} with the link function formulation in \eqref{eq:christiano}. Similar to \cite{Christiano2017DeepRL}, the utility resembles the reward function and is used to optimize the policy using a simple policy gradient RL algorithm. \cite{Gao2019PreferencebasedIM} proposed enhancements by including a better preference querying approach and a neural policy learning method with temporal differences.

\subsection{Abstractive Summarization} Summaries of this type are built by inducing the underlying ideas and concisely stating them. Its high degree of subjectivity makes it much more challenging than the extractive task. 

\break
\noindent \cite{Ziegler2019FineTuningLM} capitalized on the large capacity of pre-trained transformer models to try solving the abstractive task effectively. They use two pre-trained transformer models to represent the non-linear utility function and the underlying policy that generates summaries given the text input. The policy model is optimized using the PPO algorithm and the utility function is periodically trained with the online collection of preferences. Their summarization results turned out to be highly extractive due to the bias of the experts to prefer copied summaries and the online querying made it hard to ensure the quality of the expert responses. \cite{Stiennon2020LearningTS} solved the previous problems by shifting to a batch setting for preference collection instead of a fully online one, while consistently communicating with experts to ensure more relevant preference responses which lead to more abstractive results. \cite{Wu2021RecursivelySB} shifted the application from summarization of moderate size English text to whole books by recursively decomposing the books into smaller sections to be easily evaluated by human experts.

\section{Analysis \& Future Work}
\label{analysis}
It was seen how the integration of deep RL algorithms into the PbRL framework by \cite{Christiano2017DeepRL} revolutionized the scalability of PbRL to complex tasks and motivated a large amount of follow-up work that enhanced the efficiency successfully. Some concurrent work \cite{Pacchiano2021DuelingRR} managed to put solid foundations of theoretical guarantees for PbRL algorithms proving their robustness, while others proposed a benchmarking tool \cite{Lee2021BPrefBP} for consistent and fair comparison of PbRL algorithms. Preferences also proved their effectiveness in achieving human-desired performance in challenging real-world tasks such as text summarization \cite{Stiennon2020LearningTS}. However, open problems still exist in the current literature work along with potential future research directions.

\subsection{Formulation and performance of PbRL algorithms}State-of-the-art methods perform poorly whenever the experts make mistakes in their preferences \cite{Lee2021BPrefBP}. A possible solution could be to explicitly consider the stochasticity of the expert while designing the link functions. Also, to further enhance the feedback and sample efficiency, one could experiment with using model-based approaches like \cite{Novoseller2020DuelingPS} in more challenging environments \cite{Christiano2017DeepRL}. Additionally, not enough research has been done on handling incomparable trajectories, that could have contradicting preferences, to get Pareto-optimal policies. This could be addressed by learning a set of utility functions that are non-scalar (multi-dimensional) and utilizing multi-objective RL methods \cite{Liu2015MultiobjectiveRL} to get the corresponding Pareto-optimal policies. A limiting assumption in most work is the fact that all the trajectories in the preferences start at the same state. Such constraint could be relieved by utilizing the advantage function to represent the utility in terms of the expected rewards from the different initial states. Some methods \cite{Ibarz2018RewardLF, Stiennon2020LearningTS} utilized extra supervision signals like expert demonstrations to supplement the low amount of information gained from preferences. Other signals that could be worth experimenting with are providing explanations along with the preferences or enhancing the model output by allowing the experts to directly edit them. 

\break
\noindent Furthermore, concepts from representation learning \cite{Chen2020ASF} could be borrowed to extend PbRL to partially observable RL or sparse reward settings which require a rich representation of states.

\subsection{Safety in PbRL}To the best of our knowledge, no prior work investigated the application of PbRL in risk-averse domains. This could be implemented by over-weighing high-risk trajectories within the preferences to prioritize learning not to prefer them. In addition, further analysis should be made on adapting PbRL methods in real-world scenarios while mitigating the possibility of malicious users incurring bias in the preferences to let the model learn undesirable behaviors. 

\subsection{Theoretical Guarantees}Future work could focus on coming up with algorithms that exhibit both regret and finite-sample guarantees at the same time. Also, extending one of the approaches with regret or finite-sample guarantees to work within complex state-action spaces or infinite horizon settings can allow for their utilization in real-world domains. 

\subsection{Applications of PbRL}The state-of-the-art summarization approach by \cite{Stiennon2020LearningTS} still suffers from large feedback inefficiency. An interesting direction could integrate the pseudo-labeling approach by \cite{park2022surf} to label the existing summaries automatically without querying the expert. Moreover, there are different NLP applications involving subjective tasks which could leverage PbRL to learn human-desired behaviors and these include dialogue, machine translation, and question answering.

\section{Conclusion}
PbRL algorithms have demonstrated the possibility of utilizing human preferences as reward signals without resorting to explicit reward engineering. This survey presented the most recent advances in the field coherently while providing insights on current open problems and potential research directions. We conclude that utility-based PbRL algorithms, especially ones with non-linear formulations, provide the stepping stone to generalizing PbRL to more complex and practical application domains. However, the high cost associated with the preference feedback from experts and environment interactions creates an important research target to achieve both feedback and sample efficiency. In addition, the recent formulation of concrete regret and finite-sample guarantees initiated the tractable theoretical analysis of PbRL and future work could focus more on relaxing the assumptions of existing methods while operating under more complex environment settings. Moreover, the introduction of an open-source benchmarking tool is expected to advance the consistent evaluation of PbRL methods. Lastly, PbRL proved to reach human-desired behavior in text summarization and future work should be focused on expanding it to other subjective and challenging real-world tasks.

\bibliographystyle{IEEEtran}
\bibliography{root}

\end{document}